%% file: IEEEtran.tex
\documentclass[10pt,conference]{IEEEtran}
\IEEEoverridecommandlockouts
\usepackage[T1]{fontenc}
\input{math_commands.tex}

\usepackage{bbold}
\usepackage{cite}
\usepackage{amsmath,amssymb,amsfonts}
\usepackage{algorithmic}
\usepackage{graphicx}
\usepackage{textcomp}
\usepackage{xcolor}
\usepackage[absolute,overlay]{textpos}
\usepackage{hyperref}
\usepackage{url}
\usepackage{caption}
\usepackage{subcaption}
\usepackage[table]{xcolor}
\usepackage{booktabs}

\usepackage{float}
\usepackage{tikz}
\usepackage{wrapfig}

\usepackage{inconsolata}   % nicer monospace
\usepackage{listings}
\usepackage{enumitem}
\usepackage[most]{tcolorbox}

\lstdefinestyle{json}{
  basicstyle=\ttfamily\footnotesize,
  columns=fullflexible,
  keepspaces=true,
  breaklines=true,
  showstringspaces=false,
  upquote=true
}

\newcommand{\code}[1]{\texttt{#1}} % inline code

\usepackage[most]{tcolorbox}
\tcbuselibrary{skins,breakable}
\usepackage{xstring}
\usepackage{etoolbox}

\definecolor{darkblack}{HTML}{111111}
\definecolor{lightgray}{gray}{0.96}

\tcbuselibrary{skins,breakable}

\newtcolorbox{promptbox}[1][]{%
  enhanced,
  breakable,
  sharp corners,
  colback=lightgray!100,
  colframe=black,
  boxrule=0.4pt,
  width=\linewidth,
  left=1.5mm, right=1.5mm,
  top=1mm, bottom=1mm,
  fontupper=\scriptsize,
  title={\textbf{Prompt Template}},
  fonttitle=\bfseries\footnotesize,
  colbacktitle=black,
  coltitle=white,
  boxed title style={sharp corners, boxrule=0pt, interior style={fill=black}},
  attach boxed title to top left={xshift=30mm,yshift=-1mm},
  before upper={%
    \begingroup
    \setlength{\parindent}{0pt}%
    \setlength{\parskip}{2pt}%
    \raggedright%
    \setlist[itemize]{nosep,leftmargin=1.2em}%
    \setlist[enumerate]{nosep,leftmargin=1.6em}%
  },
  after upper=\endgroup,
  #1
}

\newtcolorbox{examplesbox}[1][]{%
  enhanced, breakable, sharp corners,
  colback=white, colframe=black,
  title={\textbf{Some Examples}},
  fonttitle=\bfseries\footnotesize,
  colbacktitle=black,
  boxed title style={size=small,colframe=black,colback=black},
  top=1mm, bottom=1mm, left=2mm, right=2mm,
  width=\linewidth,
  #1
}

\newtcolorbox{sentencebox}[1][]{
  enhanced, breakable, sharp corners,
  colback=lightgray!15,   % fundo suave
  colframe=black,         % borda preta discreta
  top=1mm, bottom=1mm, left=2mm, right=2mm,
  width=\linewidth,
  fontupper=\small,       % texto ligeiramente menor
  before skip=6pt, after skip=6pt,
  #1
}

\def\BibTeX{{\rm B\kern-.05em{\sc i\kern-.025em b}\kern-.08em
    T\kern-.1667em\lower.7ex\hbox{E}\kern-.125emX}}
\begin{document}

\title{Do Language Models Trust Their Own Justifications? A Study on Functional Consistency

\thanks{This research  was funded by Fundaç\~ao de Amparo à Pesquisa do Estado do Rio Grande do Sul (\texttt{FAPERGS}) [24/2551-0000725-3; 25/2551-0002116-2].
This work was supported by the Serrapilheira Institute (\texttt{grant number Serra - R-2501-51351}).
This study was financed in part by the Coordination for the Improvement of Higher Education Personnel – Brazil (CAPES) – Finance Code 001.}
}

\author{\IEEEauthorblockN{Alisson Rosa Pereira}
\IEEEauthorblockA{\textit{Institute of Informatics} \\
\textit{Universidade Federal do Rio Grande do Sul}\\
Porto Alegre, Brazil \\
alirpereira887@gmail.com}
\and
\IEEEauthorblockN{Bruno Iochins Grisci}
\IEEEauthorblockA{\textit{Institute of Informatics} \\
\textit{Universidade Federal do Rio Grande do Sul}\\
Porto Alegre, Brazil \\
bigrisci@inf.ufrgs.br}
}

\maketitle

\begin{textblock*}{0.85\paperwidth}
    (0.075\paperwidth, 0.92\paperheight)
    \centering
    \scriptsize

    \rule{0.4\textwidth}{0.4pt}

    \vspace{0.1cm}

    \copyright\ 2026 IEEE. Personal use of this material is permitted.
    Permission from IEEE must be obtained for all other uses,
    in any current or future media, including reprinting/republishing
    this material for advertising or promotional purposes,
    creating new collective works, for resale or redistribution
    to servers or lists, or reuse of any copyrighted component
    of this work in other works.

    \textbf{Published in:}
    A. R. Pereira and B. I. Grisci,
    ``Do Language Models Trust Their Own Justifications?
    A Study on Functional Consistency,''
    \textit{2026 International Joint Conference on Neural Networks (IJCNN)},
    Maastricht, Netherlands, 2026, pp. 1--6,
    doi: \href{https://doi.org/10.1109/IJCNN69854.2026.11728410}
    {10.1109/IJCNN69854.2026.11728410}.
\end{textblock*}

\begin{abstract}
Large Language Models (LLMs) have been widely adopted in text classification tasks, where they not only output class predictions but also generate explanations that highlight the tokens deemed most relevant for reaching the predicted label. Yet it remains unclear whether these highlighted elements are behaviorally consistent with the model’s own predictions under targeted interventions, which we refer to as functional auto-consistency. While much of the literature evaluates the textual plausibility of such explanations, few studies assess their functional consistency with the model’s actual behavior. In this work, we propose an experimental framework based on the principle of auto-consistency: if a model identifies certain tokens as decisive, then isolating, removing, or semantically inverting them should produce systematic and interpretable changes in its predictions. We operationalize this evaluation through sufficiency, comprehensiveness, and counterfactuality metrics, and conduct experiments on IMDB and Steam reviews across both closed-source (GPT-4o-mini) and open-source LLMs (Gemma3, Granite8B, DeepSeek). Results show that GPT-4o-mini follows the expected progression across all metrics, Gemma3 and Granite8B maintain coherence under sufficiency but lose consistency under more demanding interventions, while DeepSeek variants display structural deviations, either failing to preserve sufficiency or overreacting under comprehensiveness and counterfactuality. These findings show that explanation reliability varies across LLM families and scales, with smaller models displaying contradictions and larger ones exhibiting over-sensitivity. By combining sufficiency, comprehensiveness, and counterfactuality, our approach provides a systematic methodology for assessing the functional consistency of LLM self-explanations.
\end{abstract}


\section{Introduction}
\label{headings}

\textit{Large Language Models} (LLMs), built upon the \textit{Transformer} architecture \cite{vaswani2017attention}, have demonstrated remarkable abilities in generalization and task solving through textual instructions, even in scenarios where no explicit training examples are available (\textit{zero-shot} or \textit{few-shot learning})  \cite{brown2020language}. 
Among their emergent capabilities, one of the most prominent is the generation of explanations that justify predictions, often structured as reasoning chains, commonly referred to as \textit{chain-of-thought} (CoT) \cite{wei2023chainofthoughtpromptingelicitsreasoning}.

While such explanations are frequently plausible, their alignment with the model’s subsequent behavior under interventions remains an open question. 
For instance, \cite{turpin2023language} show that CoT explanations can be manipulated through biases in the prompts, thereby justifying incorrect predictions with arguments that appear convincing but are not aligned with the true decision factors. 
These findings raise concerns about excessive reliance on LLM-produced explanations and highlight the need for more objective approaches to evaluate the transparency and robustness of these models. In this work, we study functional auto-consistency: given a model’s own declared influential spans, do systematic interventions on those spans produce changes in outputs that are coherent with that declaration? Importantly, this differs from establishing ground-truth causal attribution. Our protocol is self-contained but falsifiable: if the model’s declared support does not behave as expected under interventions, the resulting contradictions are directly observable at the input–output level.

In this work, we propose an investigation that leverages \textit{feature importance} \cite{barbieri2024analysis} within the task of classification with LLMs, aiming to assess the extent to which models remain coherent with the justifications they provide for their own predictive decisions. 
This evaluation adapts the definition of self-consistency proposed by \cite{chen2023two}, understood as the requirement that model outputs remain logically non-contradictory, to what we term auto-consistency. In our setting, this notion captures whether the tokens\footnote{In this paper, the term “token” is used broadly to denote an influential text span (a word or phrase) returned in the influential terms field and extracted verbatim from the input sentence. All interventions operate on these spans, consistent with the prompt specification (“words or phrases”; see \texttt{Prompt Template}.}
 highlighted as decisive in a model’s explanation play a functionally coherent role when subjected to systematic interventions.

We hypothesize that when certain tokens are highlighted as influential for a prediction, the model should exhibit systematic variations in the assigned score when these elements are removed, isolated, or semantically modified. 
Thus, explanations are not evaluated in terms of textual plausibility or rhetorical adequacy, but rather in terms of their functional coherence with the model’s own predictive behavior.  

To quantify this coherence, we adopt the metrics of Comprehensiveness, Sufficiency, and Counterfactuality \cite{molnar2020interpretable}. 
The first two follow the definitions of \cite{deyoung2019eraser}, but are here adapted to a scalar review variable. 
Unlike probability-based formulations, we use a review score in the range $[1,10]$, predicted by the LLM, which reflects how favorably the text evaluates the target item.
\footnote{Where $1$ indicates an extremely negative evaluation and $10$ an extremely positive evaluation.} 
Under this formulation, Comprehensiveness captures the variation in the score when the highlighted tokens are removed, Sufficiency evaluates the case in which only these tokens are preserved, and Counterfactuality measures the impact of substituting them with semantically opposite terms.  

With this adaptation, we investigate whether the model behaves consistently with its own claims of importance, that is, whether it maintains functional coherence under local interventions on the tokens it designates as relevant. 
This study therefore focuses on the structural fidelity of explanations provided by LLMs, adopting a quantitative approach to the evaluation of interpretability in generative models. 
The empirical investigation is conducted in the context of sentiment classification, and the results are discussed in detail in the following sections.

\section{Related Work}

Recent work has explored LLMs as tools for feature selection, leveraging their ability to interpret textual descriptions and relate variables to predictive tasks. One line uses LLMs as sources of prior knowledge, asking models to judge feature relevance from descriptions alone (e.g., LM-Priors \cite{choi2022lmpriors} and binary prompting in \cite{brown2020language}). A second line adopts hybrid pipelines where LLMs propose feature importance but are guided to apply classical selection methods (e.g., random forests or sequential selection) to retain statistical structure \cite{li2025llm4fs}. More task-specific frameworks incorporate LLMs into iterative refinement loops (e.g., medical settings in \cite{yang2024ice}) or use them as feature engineers that generate meta-features for downstream models \cite{han2024large}.

In parallel, a growing literature studies whether LLMs can explain their own decisions through self-explanations or rationalizations \cite{madsen2024selfexplanationslargelanguagemodels,huang2023largelanguagemodelsexplain}. Empirical evidence suggests that such explanations often deviate from the factors driving predictions \cite{sarkar2024largelanguagemodelsexplain}. In the same critical vein, chain-of-thought explanations can be steered into plausible yet non-aligned narratives \cite{turpin2023language}, and models may exploit explicit cues in predictions without reflecting them in their explanations \cite{chen2025reasoningmodelsdontsay}. Collectively, these findings indicate that textual plausibility is insufficient to assess explanation reliability.

Overall, existing work converges on exploiting LLM semantics for feature relevance and explanation generation \cite{choi2022lmpriors,brown2020language,li2025llm4fs,yang2024ice,han2024large,madsen2024selfexplanationslargelanguagemodels,huang2023largelanguagemodelsexplain,turpin2023language,sarkar2024largelanguagemodelsexplain,chen2025reasoningmodelsdontsay}, yet few studies directly test whether a model’s declared evidence behaves coherently under targeted interventions. In this work, we aim to fill this gap by employing feature importance not as the central object of analysis but as a methodological mechanism for evaluating model auto-consistency. 
Specifically, we apply the metrics of Comprehensiveness, Sufficiency, and Counterfactuality to the tokens highlighted as relevant, thereby assessing whether these elements maintain functional correspondence with the model’s observed predictive behavior.

\section{Methodology}

This section outlines the methodology adopted to evaluate the functional consistency of explanations provided by LLMs in the context of classification tasks. 
We begin by describing the datasets employed and the sampling criteria used for the selection of sentences analyzed in the experiments. 
Next, we present the structure of the prompts designed to instruct the model to perform sentiment classification and to identify the tokens considered most relevant for its predictions. 
Finally, we introduce the metrics used to quantify the functional coherence of the explanations, with emphasis on the formal definitions of Comprehensiveness, Sufficiency, and Counterfactuality, which allow us to measure the extent to which the presence, absence, or semantic inversion of explanatory elements affects the model’s behavior.

\subsection{Analysis Pipeline}

The experiments were conducted using two datasets widely employed in sentiment analysis tasks. 
The first is the \textit{Stanford Large Movie Review Dataset} (IMDB) \cite{maas-EtAl:2011:ACL-HLT2011}, which consists of English-language movie reviews annotated with binary sentiment polarity (positive or negative), used here in the version available on the \textit{HuggingFace} platform.
\footnote{\url{https://huggingface.co/datasets/stanfordnlp/imdb}} 
The second dataset comprises reviews from the Steam platform \cite{pandey2022crossdomainconsumerreviewanalysis}, which collects user evaluations of PC games and enables the analysis of aspects such as player satisfaction and dissatisfaction, genre popularity, and sentiment shifts over time.

For the construction of the experimental corpus, a stratified sampling of $2.000$ sentences was performed,
ensuring proportionality between sentiment classes and thereby minimizing potential distributional biases. 
This strategy was adopted to increase representativeness and robustness in the subsequent analyses.

Model interactions were carried out through different APIs, covering recent LLM architectures: \textit{GPT-4o-mini} \cite{openai2024gpt4technicalreport}, \textit{Gemma3:4B} \cite{gemmateam2025gemma3technicalreport}, \textit{DeepSeek-R1:1.5B} and \textit{DeepSeek-R1:14B} \cite{deepseekai2025deepseekr1incentivizingreasoningcapability}\footnote{We additionally include \textit{DeepSeek-R1:14b} to enable an intra-family comparison across parameter scales. 
This choice was motivated by methodological considerations, since evaluating two models from the same provider with different capacities offers a controlled setting to assess consistency.} 
and \textit{Granite3.3:8B} \cite{mishra2024granitecodemodelsfamily}. 
The selection aimed to ensure diversity along two main axes: 
(i) proprietary versus open-source access, and 
(ii) providers from distinct research and industrial backgrounds. 
The additional intra-family comparison was conducted exclusively within the DeepSeek models, given their availability and relevance for methodological contrast.

\begin{figure}[htbp]  % posição: here, top, bottom, page
    \centering  % centraliza a imagem
    \includegraphics[width=9cm,height=4cm]{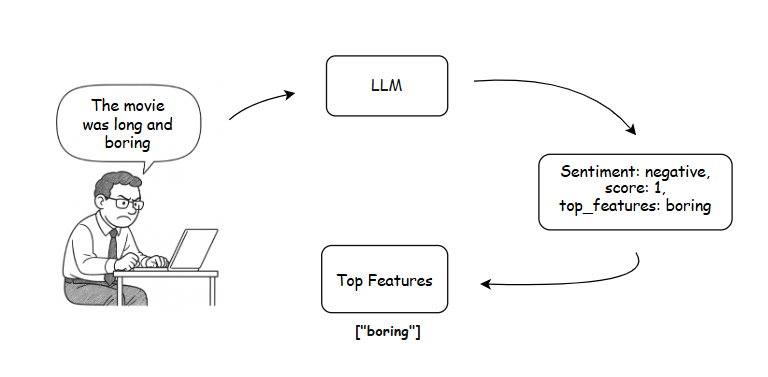}
    \caption{Sentence Analysis Workflow}
    \label{fig:diagrama}
\end{figure}

For each sentence, the model was instructed to classify the sentiment, assign a \textbf{review score} in the range $[1,10]$, and identify the tokens most relevant for the prediction. 
This score represents the intensity of the judgment produced by the model and constitutes the basis for the sufficiency, comprehensiveness, and counterfactual metrics discussed in subsequent sections. 
\autoref{fig:diagrama} illustrates this input–output pipeline.

\subsection{Prompt Formalization}

Adapting the formulation in \cite{jeong:25}, let $\mathcal{M}$ denote a pre-trained LLM. The input to $\mathcal{M}$ is a prompt $P^{\text{LLM}}$ composed of three main elements:  
\begin{itemize}
    \item a description of the dataset ($Des$);  
    \item few-shot examples ($Ex$), selected to illustrate the target task;  
    \item the task context ($C$), specifying the objective of identifying the tokens most relevant to the sentiment prediction.  
\end{itemize}

These components are combined to form the complete prompt:
\begin{equation}
    P^{\text{LLM}} = \text{prompt}(Des, Ex, C).
\end{equation}

Let $\mathcal{X} = \{x_1, x_2, \dots, x_m\}$ be a collection of sentences, where each $x_i = \{w_1^{(i)}, w_2^{(i)}, \dots, w_{n_i}^{(i)}\}$ consists of $n_i$ tokens. 
For each $x_i$, we provide $P^{\text{LLM}}$ to the model $\mathcal{M}$ together with the input sentence. 
The model then returns a subset $\mathcal{T}(x_i) \subseteq x_i$ containing the tokens identified as most relevant to justify its prediction on $x_i$:
\begin{equation}
    \mathcal{T}(x_i) = \mathcal{M}(x_i; P^{\text{LLM}}), \qquad i = 1, 2, \ldots, m.
\end{equation}

In summary, each subset $\mathcal{T}(x_i)$ corresponds to the tokens within $x_i$ that the LLM, when prompted with $P^{\text{LLM}}$, identifies as essential for supporting its sentiment classification. We instantiate $P^{\text{LLM}}$ with the following template:
\begin{promptbox}
\scriptsize
You are a sentiment analysis assistant working as a classifier for
\textbf{\{\{DOMAIN\}\}} reviews.

Your tasks are:
\begin{enumerate}
  \item Analyze the input sentence and decide \code{"positive"} or \code{"negative"}.
  \item Return \code{"sentiment"} as \code{"positive"} or \code{"negative"}.
  \item Provide a \textbf{review score} \code{"review"} from 1 to 10. Higher means stronger approval.
  \item List in \code{"influential\_terms"} the words or phrases directly extracted from the sentence that were important for the prediction.
  \item Include the original review in \code{"input\_sentence"}.
\end{enumerate}

\textbf{STRICT OUTPUT RULES}
\begin{itemize}
  \item Output must be valid JSON only.
  \item Do not include Markdown fences or explanations.
  \item Use straight quotes \code{"} and escape inner quotes with \code{\textbackslash"}.
  \item Output must start with \code{\{} and end with \code{\}}.
  \item Output must contain exactly these fields, in any order:
        \code{sentiment}, \code{review}, \code{influential\_terms}, \code{input\_sentence}.
\end{itemize}

\textbf{Examples:}
\{\{EXAMPLES\}\}
\end{promptbox}

\begin{examplesbox}
\begin{lstlisting}[style=json]
{"sentiment":"negative",
 "review":2,
 "influential_terms":["pay-to-win","constant crashes"],
 "input_sentence":"Pay-to-win mechanics, constant crashes ruined the experience for me."}
\end{lstlisting}

\begin{lstlisting}[style=json]
{"sentiment":"positive",
 "review":10,
 "influential_terms":["incredible gameplay","tight controls","runs flawlessly"],
 "input_sentence":"One of the best games I have ever played. Tight controls and it runs flawlessly even on high settings."}
\end{lstlisting}
\end{examplesbox}

\subsection{Metrics for Auto-Consistency Evaluation}

The evaluation of LLM explanations is grounded in the principle of \textbf{auto-consistency}, defined as the requirement that a model’s predictive behavior remains coherent with the set of tokens it designates as influential for its decisions. 
Formally, interventions applied to the highlighted subset $\mathcal{T} \subseteq x_i$ should induce variations in the assigned review score that are systematic and interpretable, rather than incidental.  

This principle is operationalized through three complementary metrics. 
\textit{Sufficiency} evaluates whether the tokens spans in $\mathcal{T}$, when considered in isolation, are capable of sustaining the original prediction. 
\textit{Comprehensiveness} quantifies the effect of removing $\mathcal{T}$ from the input, capturing the extent to which the prediction deteriorates in their absence. 
\textit{Counterfactuality} assesses the impact of replacing $\mathcal{T}$ with semantically opposite terms, thereby determining whether the inversion of highlighted tokens substantially alters the decision.  
Together, these metrics test whether the model’s self-reported evidence behaves coherently with its own predictions under controlled keep/remove/invert interventions.
Here’s a specific example from the Steam dataset of sentence modification:

\begin{sentencebox}

\noindent\textbf{Original Sentence:} ``Pay to win. Buggy. Doesn't deploy Hero's \\ to lanes, which is kinda the point of the whole game.''

\smallskip
\noindent\textbf{Sufficiency:} ``Pay to win. Buggy.''

\smallskip
\noindent\textbf{Comprehensiveness:} ``Doesn't deploy Hero's to lanes, which is kinda the point of the whole game.''

\smallskip
\noindent\textbf{Counterfactual Replacement:} ``Not pay to win. Stable. Doesn't deploy Hero's to lanes, which is kinda the point of the whole game.''

\end{sentencebox}

\subsubsection{Comprehensiveness}

Let $\mathcal{M}$ be a pre-trained LLM. Consider an input sequence $x_i = \{w_1, w_2, \dots, w_n\}$ consisting of $n$ tokens. 
Let $\mathcal{T} \subseteq x_i$ denote the subset of tokens highlighted by $\mathcal{M}$, when processing $x_i$ under a given prompt, as the most relevant for explaining its decision.

We denote by $\text{R}^{\mathcal{M}}(x)$ the review score assigned by model $\mathcal{M}$ to input $x$, represented as a scalar value in the range $[1,10]$. 
This review score is an ordinal signal of favorability produced by $\mathcal{M}$ under a fixed prompt. We use it comparatively within the same model (original vs. intervened inputs), without assuming cross-model calibration.

We define the \textit{comprehensiveness} metric as:

\begin{equation}
    \text{Comprehensiveness}(x_i, \mathcal{T}) = \text{R}^{\mathcal{M}}(x_i) - \text{R}^{\mathcal{M}}(x_i \setminus \mathcal{T}) \label{eq:completude}
\end{equation}

where $x_i \setminus \mathcal{T}$ denotes the input sequence with all tokens in $\mathcal{T}$ removed, and both evaluations are performed by the same model $\mathcal{M}$ on the respective texts.
\footnote{To ensure a directional interpretation consistent with the predicted label, we adopt a symmetric transformation of the review differences ($\Delta R$): when the sentence is labeled as negative, we compute $\Delta R = R_{\text{modified}} - R_{\text{original}}$; otherwise, we compute $\Delta R = R_{\text{original}} - R_{\text{modified}}$. In this way, positive values of $\Delta R$ indicate a reduction in the evaluation toward the negative pole—i.e., an adverse effect of the intervention on the model’s perception, regardless of the original label.}

\subsubsection{Sufficiency}

Within the previously established framework, we define the \textit{sufficiency} metric considering the same model $\mathcal{M}$, the input sequence $x_i$, and the subset of tokens $\mathcal{T} \subseteq x_i$ highlighted by the LLM as most relevant to its decision.

Sufficiency is given by:

\begin{equation}
    \text{Sufficiency}(x_i, \mathcal{T}) = \text{R}^{\mathcal{M}}(x_i) - \text{R}^{\mathcal{M}}(\mathcal{T}) \label{eq:suficiencia}
\end{equation}

where $\text{R}^{\mathcal{M}}(\mathcal{T})$ corresponds to the review score resulting from using only the subset $\mathcal{T}$ as input. Large differences between $\text{R}^{\mathcal{M}}(x_i)$ and $\text{R}^{\mathcal{M}}(\mathcal{T})$ indicate that the highlighted tokens, when considered in isolation, are not sufficient to sustain the original decision, thereby suggesting explanatory insufficiency.

\subsubsection{Counterfactuality}

Within the previously established framework, we define the \textit{counterfactuality} metric considering the same model $\mathcal{M}$, the input sequence $x_i$, and the subset of tokens $\mathcal{T} \subseteq x_i$ highlighted by the LLM as most relevant to its decision.

Counterfactuality is defined as:

\begin{equation}
    \text{Counterfactuality}(x_i, \mathcal{T}) = \text{R}^{\mathcal{M}}(x_i) - \text{R}^{\mathcal{M}}(x_i[\mathcal{T} \leftarrow \neg \mathcal{T}]) \label{eq:contrafactual}
\end{equation}

where $x_i[\mathcal{T} \leftarrow \neg \mathcal{T}]$ denotes the modified input sequence obtained by replacing the tokens in $\mathcal{T}$ with their semantic opposites. 

The construction of $\neg \mathcal{T}$ follows a simple and fully specified semantic-edit operator \cite{wang2024surveynaturallanguagecounterfactual}. Whenever possible, highlighted spans are replaced with antonyms retrieved from \textit{WordNet} \cite{wordnet}. When no suitable antonym exists, we apply a minimal syntactic negation heuristic by prefixing the span with ‘not’. This defines a minimal, plausible counterfactual editing rule applied uniformly across models. We acknowledge that such edits can occasionally produce unnatural phrasing; accordingly, we interpret counterfactuality as a directional stress test of dependence on $\mathcal{T}$ rather than as a guarantee of fully natural counterfactual sentences.
High values of $\text{Counterfactuality}(x_i, \mathcal{T})$ indicate stronger functional sensitivity to the highlighted spans under semantic inversion.

\subsection{Justification}

A central element of our methodology is the decision to assess auto-consistency through observable outputs rather than through confidence estimates or internal activations. Confidence/probability-based measures can be misleading in our setting because they may remain internally self-consistent under perturbations without testing whether the model’s declared supporting spans have the functional effect the model implies. Our protocol is self-contained but not tautological: $\mathcal{T}(x)$ acts as the model’s own hypothesis about support, and the intervention–re-evaluation step provides a falsification test that can reveal contradictions when outputs fail to change in the expected direction (or change erratically).

Moreover, the internal representations of proprietary LLMs are inaccessible in the inference-only setting adopted here, precluding the application of gradient-based or attention-based attribution methods. Under these constraints, the only feasible strategy is to probe consistency through behavioral interventions on the subset $\mathcal{T}(x_i)$ identified as relevant.  

Accordingly, our evaluation examines whether removing, isolating, or semantically altering $\mathcal{T}(x_i)$ produces systematic and interpretable changes in the scalar review score $R^{\mathcal{M}}(x)$. This design avoids both the circularity of probability-based measures and the opacity of internal activations, providing a rigorous and reproducible criterion for assessing the auto-consistency of LLM explanations.

\section{Evaluation Metrics}

\begin{figure}[h]   % r=à direita (l=à esquerda), largura ocupada pelo wrap
\centering
\includegraphics[
  width=52mm,
  height=50mm,
  trim=1mm 1mm 0mm 0mm, % corta bordas internas: esquerda baixo direita topo
]{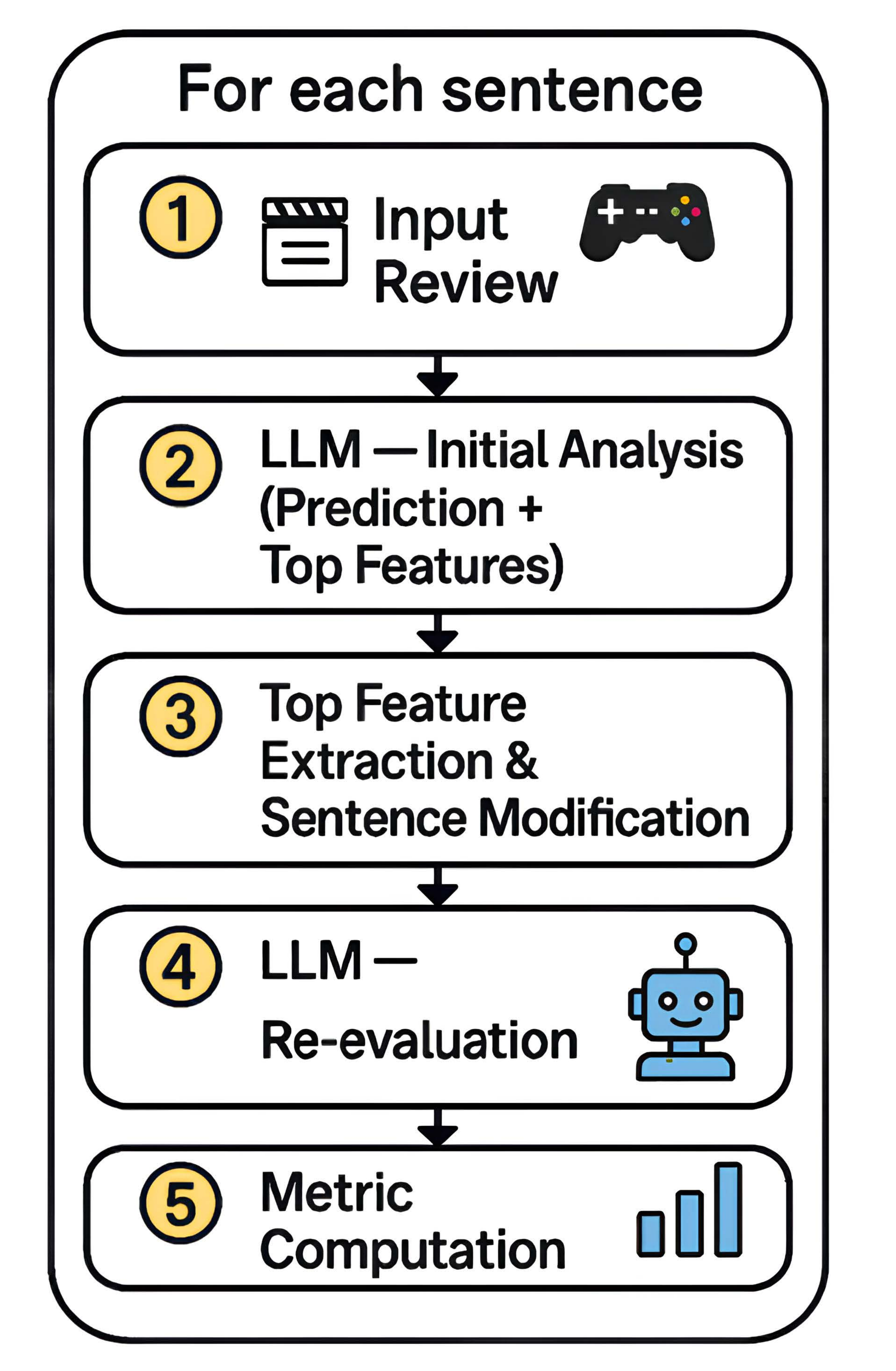}
\caption{Workflow of interventions and re-evaluations.}
\label{fig:pipeline-reeval}
\end{figure}

The experimental procedure operates at the level of individual sentences, as illustrated in Figure~\ref{fig:pipeline-reeval}. 
For each input $x_i$, the model first produces a prediction and identifies a subset of influential tokens $\mathcal{T}(x_i)$. 
Interventions are then applied to construct a modified input $x_i'$, obtained by retaining, removing, or semantically inverting $\mathcal{T}(x_i)$. 
A second evaluation by the LLM yields an updated score $\text{R}^{\mathcal{M}}(x_i')$, which can be compared against the original score $\text{R}^{\mathcal{M}}(x_i)$ to assess the functional role of the highlighted tokens.  

While this procedure provides insight into the effect of interventions at the sentence level, drawing conclusions at the scale of an entire dataset requires aggregation. 
We therefore introduce four complementary measures that summarize consistency across all sentences in a corpus:

\begin{itemize}
    \item \textbf{Mean review difference}
    ($\overline{\Delta R}$): the average of individual
    differences $\text{R}^{\mathcal{M}}(x_i) -
    \text{R}^{\mathcal{M}}(x_i')$, reflecting the overall
    effect of the intervention on the model's evaluation.

    \item \textbf{Standard deviation of differences}
    ($\sigma_{\Delta R}$): the variability of these
    differences across sentences.

    \item \textbf{Label-flip proportion}
    ($\pi_{\text{alt}}$): the fraction of instances
    where the predicted class changes after intervention.

    \item \textbf{Directional review reduction}
    ($\pi_{\text{red}}$): the proportion of cases in which
    the intervention drives the score toward the opposite
    sentiment polarity. Formally, let
    $\hat{y}^{\mathcal{M}}(x_i) \in
    \{\text{positive}, \text{negative}\}$
    denote the label predicted for the original sentence.
\end{itemize}

Then:
\[
\Delta R_i =
\begin{cases}
    \text{R}^{\mathcal{M}}(x_i) -
    \text{R}^{\mathcal{M}}(x_i'),
    & \text{if } \hat{y}^{\mathcal{M}}(x_i) = \text{positive},\\
    \text{R}^{\mathcal{M}}(x_i') -
    \text{R}^{\mathcal{M}}(x_i),
    & \text{if } \hat{y}^{\mathcal{M}}(x_i) = \text{negative}.
\end{cases}
\]

\noindent
We compute
$\pi_{\text{red}} =
\frac{1}{m}\sum_{i=1}^{m}
\mathbb{1}[\Delta R_i > 0]$.

\section{Experimental Results}

\autoref{tab:self_consistency_summary} reports the outcomes of the sufficiency, comprehensiveness, and counterfactuality experiments across IMDB and Steam.
A consistent trajectory emerges for Gemma3:4B, GPT-4o-mini, and Granite8B: highlighted tokens generally preserve the original decision when isolated, their removal weakens predictions without systematically overturning them, and semantic inversion produces stronger shifts, as expected when the meaning of decisive elements is reversed.

The DeepSeek models deviate from this trajectory.
The 1.5B variant shows instability from the outset, with elevated flip rates under sufficiency that persist across the other interventions.
The 14B variant, in contrast, maintains greater stability in sufficiency but reacts disproportionately when tokens are removed or inverted, leading to shifts and class changes substantially above those observed in the other models.

\begin{table}[t]
\centering
\caption{Summary of auto-consistency metrics on the \textit{IMDB} and \textit{Steam} datasets. Selected results are discussed in \autoref{sec:discussion}.}
\label{tab:self_consistency_summary}
\scriptsize
\resizebox{\linewidth}{!}{
\begin{tabular}{lcccccccc}
\toprule
 & \multicolumn{4}{c}{IMDB} & \multicolumn{4}{c}{Steam} \\
\cmidrule(lr){2-5} \cmidrule(lr){6-9}
Model & $\overline{\Delta R}$ & $\sigma_{\Delta R}$ & $\pi_{\text{alt}}$ [\%] & $\pi_{\text{red}}$ [\%] & $\overline{\Delta R}$ & $\sigma_{\Delta R}$ & $\pi_{\text{alt}}$ [\%] & $\pi_{\text{red}}$ [\%] \\
\midrule
\multicolumn{9}{c}{Sufficiency} \\
\midrule
Gemma3:4B     & 0.02  & 1.48 & 3.8  & 14.2 & 0.27  & 1.84 & 6.9  & 16.4 \\
GPT-4o-mini   & -0.28 & 0.87 & 0.9  & 10.5 & -0.09 & 1.33 & 3.5  & 13.3 \\
Granite8B     & -0.41 & 1.42 & 3.8  &  9.9 & 0.13  & 1.89 & 6.7  & 21.0 \\
DeepSeek-1.5B & 0.66  & 2.84 & 29.1 & 43.6 & 0.56  & 2.91 & 30.9 & 42.9 \\
DeepSeek-14B  & -0.26 & 1.51 & 13.7 & 21.0 & -0.02 & 1.75 & 10.6 & 22.3 \\
\midrule
\multicolumn{9}{c}{Comprehensiveness} \\
\midrule
Gemma3:4B     & 0.91  & 1.96 & 15.7 & 32.1 & 1.74  & 2.82 & 24.5 & 44.3 \\
GPT-4o-mini   & 1.21  & 1.70 & 19.9 & 53.9 & 1.80  & 2.71 & 28.5 & 50.0 \\
Granite8B     & 0.54  & 1.21 &  6.1 & 35.1 & 1.54  & 2.62 & 20.3 & 48.0 \\
DeepSeek-1.5B & 0.55  & 2.66 & 39.0 & 38.9 & 0.89  & 2.88 & 39.4 & 44.3 \\
DeepSeek-14B  & 0.62  & 1.42 & 31.5 & 44.2 & 1.68  & 2.76 & 45.5 & 45.5 \\
\midrule
\multicolumn{9}{c}{Counterfactuality} \\
\midrule
Gemma3:4B     & 2.04  & 2.76 & 32.4 & 49.7 & 2.71  & 3.25 & 41.5 & 50.5 \\
GPT-4o-mini   & 2.43  & 2.41 & 43.6 & 65.9 & 2.28  & 2.72 & 38.3 & 57.0 \\
Granite8B     & 1.50  & 2.20 & 24.3 & 53.5 & 2.36  & 3.14 & 33.9 & 54.4 \\
DeepSeek-1.5B & 0.89  & 2.51 & 45.4 & 45.9 & 1.16  & 2.87 & 44.9 & 44.9 \\
DeepSeek-14B  & 1.90  & 2.20 & 49.4 & 60.3 & 2.65  & 3.08 & 49.4 & 49.4 \\
\bottomrule
\end{tabular}
}
\end{table}

Taken together, the results distinguish two profiles, 
Gemma3, GPT-4o-mini, and Granite8B follow the expected progression of increasing sensitivity across interventions, whereas the DeepSeek models diverge—one by failing to preserve stability under sufficiency, the other by exhibiting excessive volatility under stronger perturbations.

\section{Discussion} \label{sec:discussion}

Our results show that highlighted influential terms have \emph{model-dependent} functional meaning under targeted interventions. Under the same keep/remove/invert protocol, LLMs fall into distinct auto-consistency regimes that only become clear when sufficiency, comprehensiveness, and counterfactuality are analyzed jointly (\autoref{tab:self_consistency_summary}). Therefore, highlighted spans are better treated as a hypothesis about decision support than as feature importance.

This also clarifies why self-explanation is a fragile attribution mechanism. Sufficiency and comprehensiveness probe different roles: the highlighted subset $\mathcal{T}$ may be \emph{sufficient} to preserve the decision in isolation while not being \emph{necessary} when removed, because sentiment evidence is distributed across the sentence. In our results this appears as a large gap between sufficiency and removal flips (e.g., GPT-4o-mini: IMDB $\pi_{\text{alt}}$ $0.9\%\rightarrow 19.9\%$, Steam $3.5\%\rightarrow 28.5\%$). Additionally, sufficiency can push scores in the opposite direction of a strict ``importance'' reading (GPT-4o-mini has negative $\overline{\Delta R}$ under sufficiency on both datasets), consistent with cases where isolating highlighted spans removes conflicting context and \emph{increases} apparent confidence.

GPT-4o-mini still follows the expected ordering of intervention strength, with inversion producing the largest disruption (IMDB $\pi_{\text{alt}}=43.6\%$, Steam $38.3\%$), while Gemma3:4B and Granite8B show a similar regime with low instability under sufficiency (IMDB $\pi_{\text{alt}}=3.8\%$ for both), suggesting highlighted spans act as meaningful but often redundant cues. DeepSeek deviates in two ways: DeepSeek-1.5B is unstable already under sufficiency (IMDB $29.1\%$, Steam $30.9\%$), whereas DeepSeek-14B is more stable under sufficiency (IMDB $13.7\%$, Steam $10.6\%$) but highly sensitive under removal (Steam comprehensiveness $\pi_{\text{alt}}=45.5\%$).

Overall, these findings caution against using LLM self-reported spans as feature importance without behavioral validation: the highlighted subset can be sufficient yet not necessary, and it can even yield higher apparent confidence when isolated. The coupled metrics provide a compact falsification test that separates informative-but-partial evidence from instability under isolation or over-sensitivity under stronger perturbations.

\subsection{Limitations and Future Work}

We evaluate only two sentiment datasets (IMDB and Steam), so conclusions may not transfer to domains with different linguistic properties. 
Our interventions operate at the token/span level and therefore do not capture higher-level structure (e.g., discourse or compositional effects). 
Token extraction itself can introduce noise, particularly for smaller models, which may affect metric stability. 
Counterfactual edits rely on a heuristic operator (WordNet antonyms or negation) and may yield unnatural phrasing, so counterfactuality should be interpreted as a directional stress test rather than a semantically perfect counterfactual. 
In addition, we position this study as a diagnostic protocol; therefore we do not benchmark against external attribution baselines (e.g., length-matched random spans), and all results are obtained under a single prompt template.

Future work should broaden coverage to additional datasets and tasks, add span-agnostic controls (e.g., random or length-matched spans), refine counterfactual generation with richer semantic resources, and study sensitivity to alternative prompt formulations. 
Combining behavioral metrics with human-centered evaluation would also clarify when auto-consistency correlates with explanation usefulness in practice.

%If you'd like to, you may include  a section for author contributions as is done in many journals. This is optional and at the discretion of the authors.
%The contributions will be included after the reviewing process.

\subsection*{Reproducibility Statement}

All experiments were run in Python 3.13 on Ubuntu 24.04.3 LTS. The codebase and prompts will be released after the review process. We use IMDB and Steam reviews, and evaluate open-source LLMs (Gemma3, Granite8B, DeepSeek 1.5B/14B via Ollama) plus GPT-4o-mini via the OpenAI API. This setup supports replication and extensions to other datasets, models, and prompt variants.

\subsection*{Acknowledgments}
This research  was funded by Fundaç\~ao de Amparo à Pesquisa do Estado do Rio Grande do Sul (\texttt{FAPERGS}) [24/2551-0000725-3; 25/2551-0002116-2].
This work was supported by the Serrapilheira Institute (\texttt{grant number Serra - R-2501-51351}).
This study was financed in part by the Coordination for the Improvement of Higher Education Personnel – Brazil (CAPES) – Finance Code 001.

\bibliographystyle{IEEEtran}
\bibliography{references}

\end{document}

%% file: math_commands.tex
\usepackage{amsmath,amsfonts,bm}

\def\eqref#1{equation~\ref{#1}}
\def\1{\bm{1}}

\DeclareMathAlphabet{\mathsfit}{\encodingdefault}{\sfdefault}{m}{sl}
\SetMathAlphabet{\mathsfit}{bold}{\encodingdefault}{\sfdefault}{bx}{n}

%% file: IEEEtran.bbl
\begin{thebibliography}{10}
\providecommand{\url}[1]{#1}
\csname url@samestyle\endcsname
\providecommand{\newblock}{\relax}
\providecommand{\bibinfo}[2]{#2}
\providecommand{\BIBentrySTDinterwordspacing}{\spaceskip=0pt\relax}
\providecommand{\BIBentryALTinterwordstretchfactor}{4}
\providecommand{\BIBentryALTinterwordspacing}{\spaceskip=\fontdimen2\font plus
\BIBentryALTinterwordstretchfactor\fontdimen3\font minus \fontdimen4\font\relax}
\providecommand{\BIBforeignlanguage}[2]{{%
\expandafter\ifx\csname l@#1\endcsname\relax
\typeout{** WARNING: IEEEtran.bst: No hyphenation pattern has been}%
\typeout{** loaded for the language `#1'. Using the pattern for}%
\typeout{** the default language instead.}%
\else
\language=\csname l@#1\endcsname
\fi
#2}}
\providecommand{\BIBdecl}{\relax}
\BIBdecl

\bibitem{vaswani2017attention}
A.~Vaswani, N.~Shazeer, N.~Parmar, J.~Uszkoreit, L.~Jones, A.~N. Gomez, {\L}.~Kaiser, and I.~Polosukhin, ``Attention is all you need,'' \emph{Advances in neural information processing systems}, vol.~30, 2017.

\bibitem{brown2020language}
T.~Brown, B.~Mann, N.~Ryder, M.~Subbiah, J.~D. Kaplan, P.~Dhariwal, A.~Neelakantan \emph{et~al.}, ``Language models are few-shot learners,'' \emph{Advances in neural information processing systems}, vol.~33, pp. 1877--1901, 2020.

\bibitem{wei2023chainofthoughtpromptingelicitsreasoning}
\BIBentryALTinterwordspacing
J.~Wei, X.~Wang, D.~Schuurmans, M.~Bosma, B.~Ichter, F.~Xia, E.~Chi, Q.~Le, and D.~Zhou, ``Chain-of-thought prompting elicits reasoning in large language models,'' 2023. [Online]. Available: \url{https://arxiv.org/abs/2201.11903}
\BIBentrySTDinterwordspacing

\bibitem{turpin2023language}
M.~Turpin, J.~Michael, E.~Perez, and S.~Bowman, ``Language models don't always say what they think: Unfaithful explanations in chain-of-thought prompting,'' \emph{Advances in Neural Information Processing Systems}, vol.~36, pp. 74\,952--74\,965, 2023.

\bibitem{barbieri2024analysis}
M.~C. Barbieri, B.~I. Grisci, and M.~Dorn, ``Analysis and comparison of feature selection methods towards performance and stability,'' \emph{Expert Systems with Applications}, vol. 249, p. 123667, 2024.

\bibitem{chen2023two}
A.~Chen, J.~Phang, A.~Parrish, V.~Padmakumar, C.~Zhao, S.~R. Bowman, and K.~Cho, ``Two failures of self-consistency in the multi-step reasoning of llms,'' \emph{arXiv preprint arXiv:2305.14279}, 2023.

\bibitem{molnar2020interpretable}
C.~Molnar, \emph{Interpretable machine learning}.\hskip 1em plus 0.5em minus 0.4em\relax Lulu. com, 2020.

\bibitem{deyoung2019eraser}
J.~DeYoung, S.~Jain, N.~F. Rajani, E.~Lehman, C.~Xiong, R.~Socher, and B.~C. Wallace, ``Eraser: A benchmark to evaluate rationalized nlp models,'' \emph{arXiv preprint arXiv:1911.03429}, 2019.

\bibitem{choi2022lmpriors}
K.~Choi, C.~Cundy, S.~Srivastava, and S.~Ermon, ``Lmpriors: Pre-trained language models as task-specific priors,'' \emph{arXiv preprint arXiv:2210.12530}, 2022.

\bibitem{li2025llm4fs}
J.~Li and X.~Xiu, ``Llm4fs: Leveraging large language models for feature selection and how to improve it,'' \emph{arXiv preprint arXiv:2503.24157}, 2025.

\bibitem{yang2024ice}
T.~Yang, T.~Yang, F.~Lyu, S.~Liu \emph{et~al.}, ``Ice-search: A language model-driven feature selection approach,'' \emph{arXiv preprint arXiv:2402.18609}, 2024.

\bibitem{han2024large}
S.~Han, J.~Yoon, S.~O. Arik, and T.~Pfister, ``Large language models can automatically engineer features for few-shot tabular learning,'' \emph{arXiv preprint arXiv:2404.09491}, 2024.

\bibitem{madsen2024selfexplanationslargelanguagemodels}
\BIBentryALTinterwordspacing
A.~Madsen, S.~Chandar, and S.~Reddy, ``Are self-explanations from large language models faithful?'' 2024. [Online]. Available: \url{https://arxiv.org/abs/2401.07927}
\BIBentrySTDinterwordspacing

\bibitem{huang2023largelanguagemodelsexplain}
\BIBentryALTinterwordspacing
S.~Huang, S.~Mamidanna, S.~Jangam, Y.~Zhou, and L.~H. Gilpin, ``Can large language models explain themselves? a study of llm-generated self-explanations,'' 2023. [Online]. Available: \url{https://arxiv.org/abs/2310.11207}
\BIBentrySTDinterwordspacing

\bibitem{sarkar2024largelanguagemodelsexplain}
\BIBentryALTinterwordspacing
A.~Sarkar, ``Large language models cannot explain themselves,'' 2024. [Online]. Available: \url{https://arxiv.org/abs/2405.04382}
\BIBentrySTDinterwordspacing

\bibitem{chen2025reasoningmodelsdontsay}
\BIBentryALTinterwordspacing
Y.~Chen, J.~Benton, A.~Radhakrishnan, J.~Uesato, C.~Denison, J.~Schulman \emph{et~al.}, ``Reasoning models don't always say what they think,'' 2025. [Online]. Available: \url{https://arxiv.org/abs/2505.05410}
\BIBentrySTDinterwordspacing

\bibitem{maas-EtAl:2011:ACL-HLT2011}
\BIBentryALTinterwordspacing
A.~L. Maas, R.~E. Daly, P.~T. Pham, D.~Huang, A.~Y. Ng, and C.~Potts, ``Learning word vectors for sentiment analysis,'' in \emph{Proceedings of the 49th Annual Meeting of the Association for Computational Linguistics: Human Language Technologies}.\hskip 1em plus 0.5em minus 0.4em\relax Portland, Oregon, USA: Association for Computational Linguistics, June 2011, pp. 142--150. [Online]. Available: \url{http://www.aclweb.org/anthology/P11-1015}
\BIBentrySTDinterwordspacing

\bibitem{pandey2022crossdomainconsumerreviewanalysis}
\BIBentryALTinterwordspacing
A.~Pandey and K.~Joshi, ``Cross-domain consumer review analysis,'' 2022. [Online]. Available: \url{https://arxiv.org/abs/2212.13916}
\BIBentrySTDinterwordspacing

\bibitem{openai2024gpt4technicalreport}
\BIBentryALTinterwordspacing
OpenAI, J.~Achiam, S.~Adler, S.~Agarwal, L.~Ahmad, I.~Akkaya, F.~L. Aleman \emph{et~al.}, ``Gpt-4 technical report,'' 2024. [Online]. Available: \url{https://arxiv.org/abs/2303.08774}
\BIBentrySTDinterwordspacing

\bibitem{gemmateam2025gemma3technicalreport}
\BIBentryALTinterwordspacing
G.~Team, A.~Kamath, J.~Ferret, S.~Pathak, N.~Vieillard, R.~Merhej, S.~Perrin \emph{et~al.}, ``Gemma 3 technical report,'' 2025. [Online]. Available: \url{https://arxiv.org/abs/2503.19786}
\BIBentrySTDinterwordspacing

\bibitem{deepseekai2025deepseekr1incentivizingreasoningcapability}
\BIBentryALTinterwordspacing
DeepSeek-AI, D.~Guo, D.~Yang, H.~Zhang, J.~Song, R.~Zhang, R.~Xu \emph{et~al.}, ``Deepseek-r1: Incentivizing reasoning capability in llms via reinforcement learning,'' 2025. [Online]. Available: \url{https://arxiv.org/abs/2501.12948}
\BIBentrySTDinterwordspacing

\bibitem{mishra2024granitecodemodelsfamily}
M.~Mishra, M.~Stallone, G.~Zhang, Y.~Shen, A.~Prasad, A.~M. Soria \emph{et~al.}, ``Granite code models: A family of open foundation models for code intelligence,'' 2024.

\bibitem{jeong:25}
D.~P. Jeong, Z.~C. Lipton, and P.~Ravikumar, ``Llm-select: Feature selection with large language models,'' in \emph{arXiv preprint arXiv:2407.02694}, arXiv Authors, Ed.\hskip 1em plus 0.5em minus 0.4em\relax arXiv, 2025, pp. 1--77.

\bibitem{wang2024surveynaturallanguagecounterfactual}
\BIBentryALTinterwordspacing
Y.~Wang, X.~Qiu, Y.~Yue, X.~Guo, Z.~Zeng, Y.~Feng, and Z.~Shen, ``A survey on natural language counterfactual generation,'' 2024. [Online]. Available: \url{https://arxiv.org/abs/2407.03993}
\BIBentrySTDinterwordspacing

\bibitem{wordnet}
C.~Fellbaum, ``Wordnet,'' in \emph{Theory and applications of ontology: computer applications}.\hskip 1em plus 0.5em minus 0.4em\relax Springer, 2010, pp. 231--243.

\end{thebibliography}
